\documentclass[runningheads]{llncs}
\usepackage[T1]{fontenc}
\usepackage{amsmath} 
\usepackage{multirow}
\usepackage{booktabs}

\usepackage{graphicx}
\begin{document}
\title{SLearnLLM: A Self-Learning Framework for Efficient Domain-Specific Adaptation of Large Language Models}
\titlerunning{Abbreviated paper title}


\author{Xiang Liu$\dag$\inst{1,2} \and
Zhaoxiang Liu$\dag$*\inst{1,2} \and
Peng Wang\inst{3} \and
Kohou Wang\inst{1,2} \and
Huan Hu\inst{1,2} \and
Kai Wang\inst{1,2} \and
Shiguo Lian*\inst{1,2}
}

\authorrunning{J. Huang et al.}
%
\institute
{Unicom Data Intelligence, China Unicom \and
Data Science \& Artificial Intelligence Research Institute, China Unicom \and
China United Network Communications Group Corporation Limited 
\email
{
\{liux750,liuzx178,wangpeng,wangzp103,huh30,wangk115,liansg\}@chinaunicom.cn 
$\dag$Equal contribution,*Corresponding author(s)
}
}
%
\maketitle              
\begin{abstract}
	When using supervised fine-tuning (SFT) to adapt large language models (LLMs) to specific domains, a significant challenge arises: should we use the entire SFT dataset for fine-tuning? Common practice often involves fine-tuning directly on the entire dataset due to limited information on the LLM's past training data. However, if the SFT dataset largely overlaps with the model's existing knowledge, the performance gains are minimal, leading to wasted computational resources. Identifying the unknown knowledge within the SFT dataset and using it to fine-tune the model could substantially improve the training efficiency. To address this challenge, we propose a self-learning framework for LLMs inspired by human learning pattern. This framework takes a fine-tuning (SFT) dataset in a specific domain as input. First, the LLMs answer the questions in the SFT dataset. The LLMs then objectively grade the responses and filter out the incorrectly answered QA pairs. Finally, we fine-tune the LLMs based on this filtered QA set. Experimental results in the fields of agriculture and medicine demonstrate that our method substantially reduces training time while achieving comparable improvements to those attained with full dataset fine-tuning. By concentrating on the unknown knowledge within the SFT dataset, our approach enhances the efficiency of fine-tuning LLMs.

\end{abstract}

\section{Introduction}\label{Indro}

Large language models (LLMs)~\cite{achiam2023gpt,anil2023palm,team2023gemini,bai2023qwen,touvron2023llama,touvron2023llamaV2}, such as GPT-4 and Claude 2, have achieved remarkable breakthroughs and progress across a wide array of domains. They have demonstrated significant potential in general areas like natural language understanding, dialogue systems, text generation, and composition. 
However, in real-world applications, the demand for LLMs in specialized domains is even greater. While LLMs have already performed well in many specialized domains, there is still a gap in achieving the high precision required for production use. If LLMs yield unsatisfactory results or even generate hallucinations~\cite{zhang2023siren}, the impact could be immeasurable.

Supervised fine-tuning (SFT)~\cite{ouyang2022training} is a widely used method to enhance a model's performance to a production-ready level in specific domains. However, when LLMs perform only moderately well, the SFT dataset often largely overlaps with the model's existing knowledge. This overlap makes the fine-tuning process inefficient, as learning already acquired knowledge consumes computational resources without contributing to any meaningful improvement.

\begin{figure*}[t!]
    \centering
    \includegraphics[width=\textwidth]{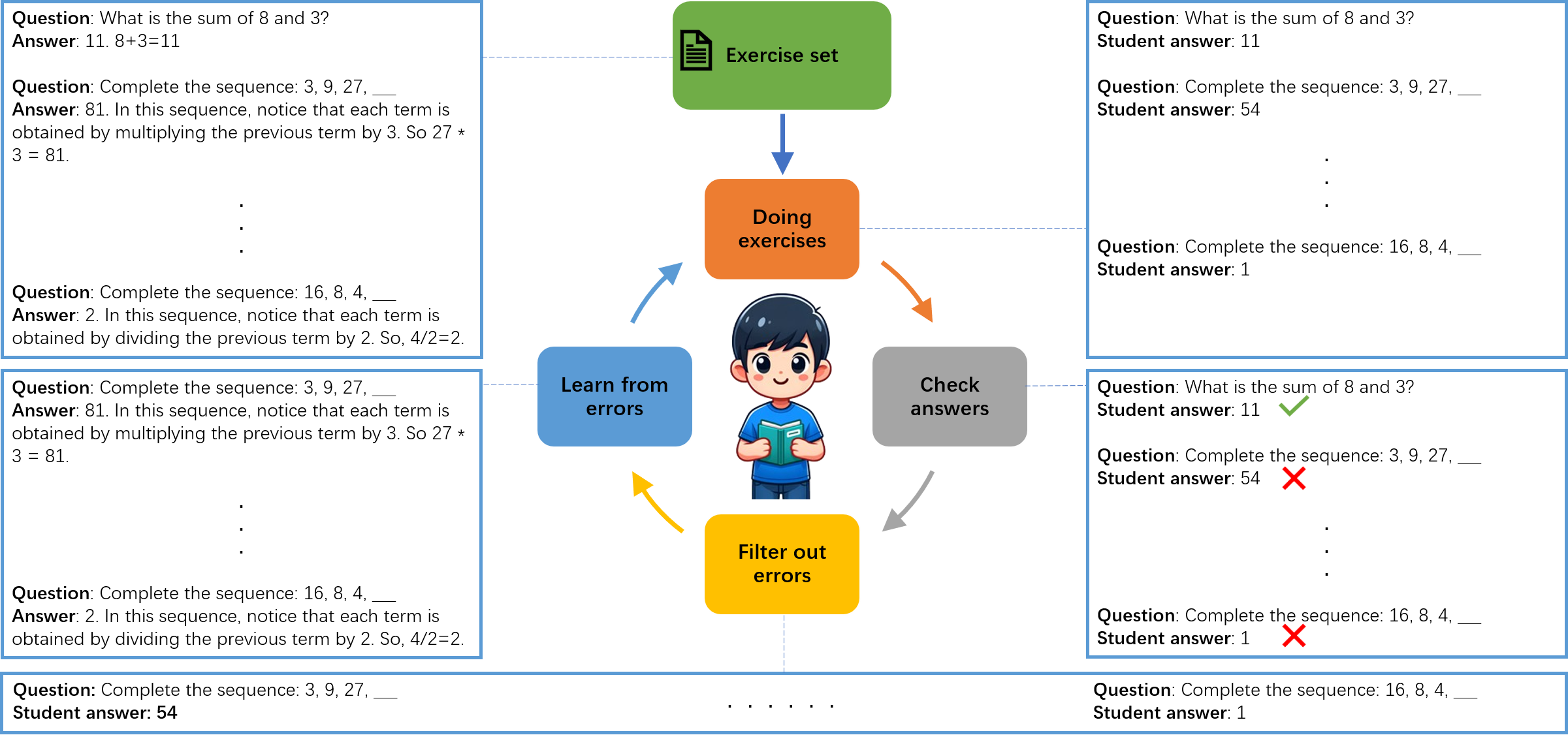}
    \caption{An efficient human learning pattern that focuses on learning unknown knowledge. This pattern facilitates knowledge acquisition and consolidation through a continuous learning cycle. In this cycle, students undergo several key steps: doing exercises, check answers, filter out errors and learn from errors. Through this iterative process, students gradually master the knowledge in the exercise set.} \label{fig1}
\end{figure*}

Now, let's turn to an efficient human learning pattern that focuses on learning unknown knowledge. As depicted in Figure~\ref{fig1}, it illustrates a widely recognized human learning pattern that facilitates knowledge acquisition and consolidation through a continuous learning cycle. In this cycle, students undergo several key steps: they receive an exercise set and answer the questions independently; they then check their answers against the reference answers; they carefully filter out the questions they answered incorrectly or did not fully understand; and finally, they focus on analyzing and learning the knowledge points behind these incorrect-answered questions to enhance their abilities. Through this iterative process, students gradually master the knowledge in the exercise set.

Drawing inspiration from human learning methods, we proposed an efficient self-learning framework for LLMs (SLearnLLM). This framework takes a SFT dataset in a specific domain as input. LLMs are utilized to answer the questions in the SFT dataset. Afterward, the LLMs, like students self-checking their own work, grade their responses by comparing them to the reference answers. Based on these scores, the LLMs assess and filter out the QA pairs with incorrect answers. Finally, the LLMs are fine-tuned using this set of incorrectly answered QA pairs.
Through SLearnLLM, the unknown knowledge within the specific domain corpus can be pinpointed, and LLMs can be fine-tuned using this knowledge, thereby accelerating the learning process and reducing computational costs.

The contributions of our paper can be summarized as follows:
\begin{itemize}
    \item The self-learning framework for LLMs: Our framework targets LLMs with strong logical reasoning and deep contextual understanding~\cite{chang2024survey}, as these models, like students, possess self-learning abilities. Models with parameters exceeding 6 billion, such as LLaMA2-7B, LLaMA-13B, Qwen1.5-7B, and Qwen1.5-14B, etc., are particularly suited for this framework. By using the framework, we can significantly enhance the performance of these models in specific domains with remarkable efficiency.
    
    
    \item The method for identifying unknown knowledge: In our framework, we introduce an effective method for identifying unknown knowledge in LLMs within specific domains.

    \end{itemize}

\section{Related work}
In recent years, there has been a surge of interest in exploring the self-learning framework of LLMs. Several notable approaches have emerged that aim to improve the performance of LLMs through various self-improving mechanisms.

Ferdinan~\cite{ferdinan2024into} proposed a self-learning framework that guides LLMs to generate questions based on a given topic, checks the consistency (variance)~\cite{manakul2023selfcheckgpt} of multiple generated responses to identify potential hallucinations, and retrieves more accurate answers from external knowledge sources for hallucinated questions and uses these data to fine-tune the LLM. 
Its method utilizing the consistency of LLM's multiple responses to evaluate unknown knowledge, might be feasible in our scenario, however, it is quite inefficient when dealing with a corpus of a specific domain.

Huang~\cite{huang2022large} utilized a pre-trained LLM to generate "high-confidence" rationale-augmented answers for unlabeled questions. This was achieved through the use of Chain-of-Thought (CoT) prompting~\cite{wei2022chain} and self-consistency~\cite{wang2022self} techniques. The LLM was then fine-tuned using these self-generated solutions as target outputs.
Tian~\cite{tian2024toward} introduced AlphaLLM, a framework that integrates Monte Carlo Tree Search~\cite{de2016monte} with LLMs to establish a self-improving loop. Using the initial dataset of expert-generated prompt-response pairs, AlphaLLM iteratively guides the model to generate subsequent prompt-response data while simultaneously updating the model. This approach enhances the capabilities of LLMs without requiring additional annotations. 

Singh~\cite{singh2023beyond} focuses on improving language models' problem-solving capabilities by self-training on a large-scale, synthetic dataset. This approach is designed to surpass the limitations of human-labeled data by generating high-quality, diverse problem-solving data through the model itself. The authors demonstrate that self-training can enhance model performance across various reasoning tasks, pushing forward the scalability and adaptability of language models in complex problem-solving scenarios.

InfuserKI~\cite{wang2024infuserki} leverages Knowledge Graphs (KGs) to enhance LLMs in knowledge-intensive tasks. It transforms knowledge triplets into multiple-choice questions and statements using GPT-4-generated templates and extracts answers from LLM outputs with regular expressions. InfuserKI employs three training phases to integrate structured KG information. While InfuserKI is effective for multiple-choice datasets, its reliance on regular expressions limits its applicability to non-multiple-choice datasets and reduces its adaptability for domain-specific SFT datasets.

Above all, These methods generate training data based on the initial dataset or question set without involving the filtering and selection of unknown knowledge. In contrast, our approach focuses on identifying questions within the QA pairs converted from a specific domain corpus that the model answers incorrectly. We then use these selected questions and their corresponding answers to fine-tune the model. Our primary objective is to achieve more efficient, targeted learning from a large corpus. By doing so, we ensure the model's improvements are directly related to its weaknesses, resulting in a more efficient enhancement of its capabilities within the specific domain. 

\section{Method}
We will present the overall process of our proposed framework, then introduce the key steps in detail, including scoring the answers and filtering out incorrect-answered questions.

\subsection{Self-learning framework for LLMs}  
Our primary objective with SLearnLLM is to efficiently leverage the given SFT dataset from a specific domain to automatically enhance the performance of the target LLM in that specific domain.

\begin{figure}[t!]
    \centering
    \includegraphics[height=8cm]{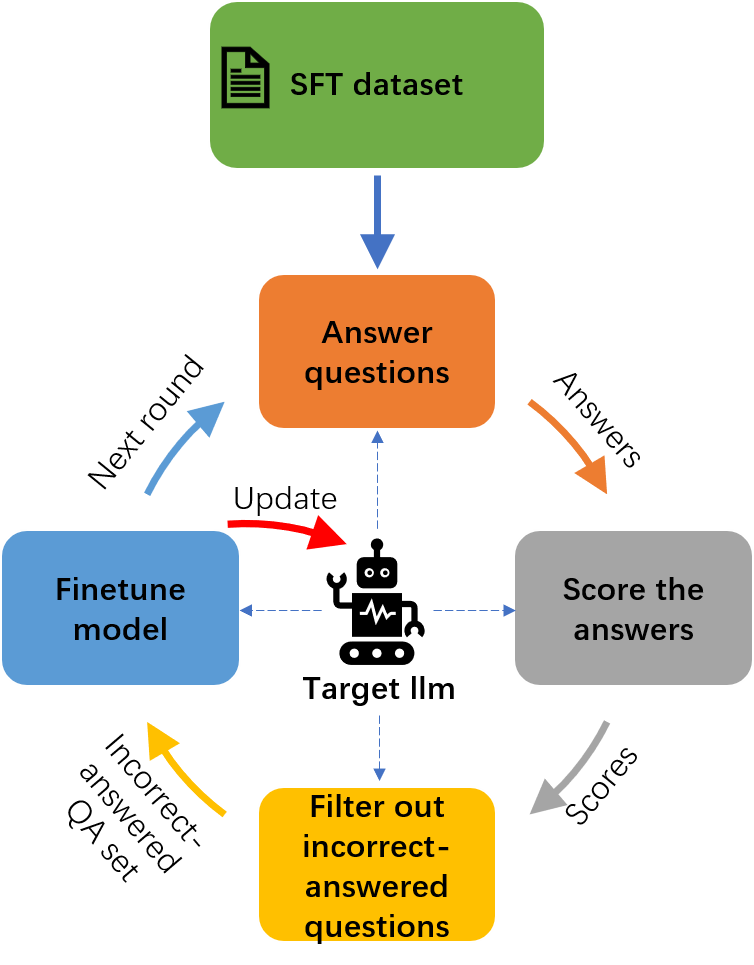}
    \caption{The framework of SLearnLLM. The framework comprises four steps: answering questions, scoring responses against the QA sets using the target LLM with robust logical reasoning, filtering out incorrect-answered questions, and fine-tuning the model based on incorrect-answered QA set to enhance performance in specific domains.} \label{fig2}
\end{figure}

Our SLearnLLM comprises four steps. Figure~\ref{fig2} illustrates the framework for the target LLM.

1. Answer questions: The target LLM responds to the questions in the SFT dataset.

2. Score the answers: The responses are compared to the answers in the SFT dataset, and scores are assigned by the target LLM based on its logical reasoning capabilities.

3. Filter out incorrect-answered questions: Based on the scores, the target LLM identifies the QA pairs where the responses are incorrect, creating a new dataset, which we refer to as the incorrect-answered QA set.

4. Fine-tune the model: Utilizing SFT, the target LLM is refined based on the incorrect-answered QA set, enhancing its performance within the specific domain.

Through these four steps, we leverage the logical reasoning capabilities of LLMs to filter out the incorrect-answered SFT dataset. Fine-tuning LLMs on this dataset can significantly improve training efficiency.

\begin{figure*}[t!]
    \centering
    \includegraphics[width=\textwidth]{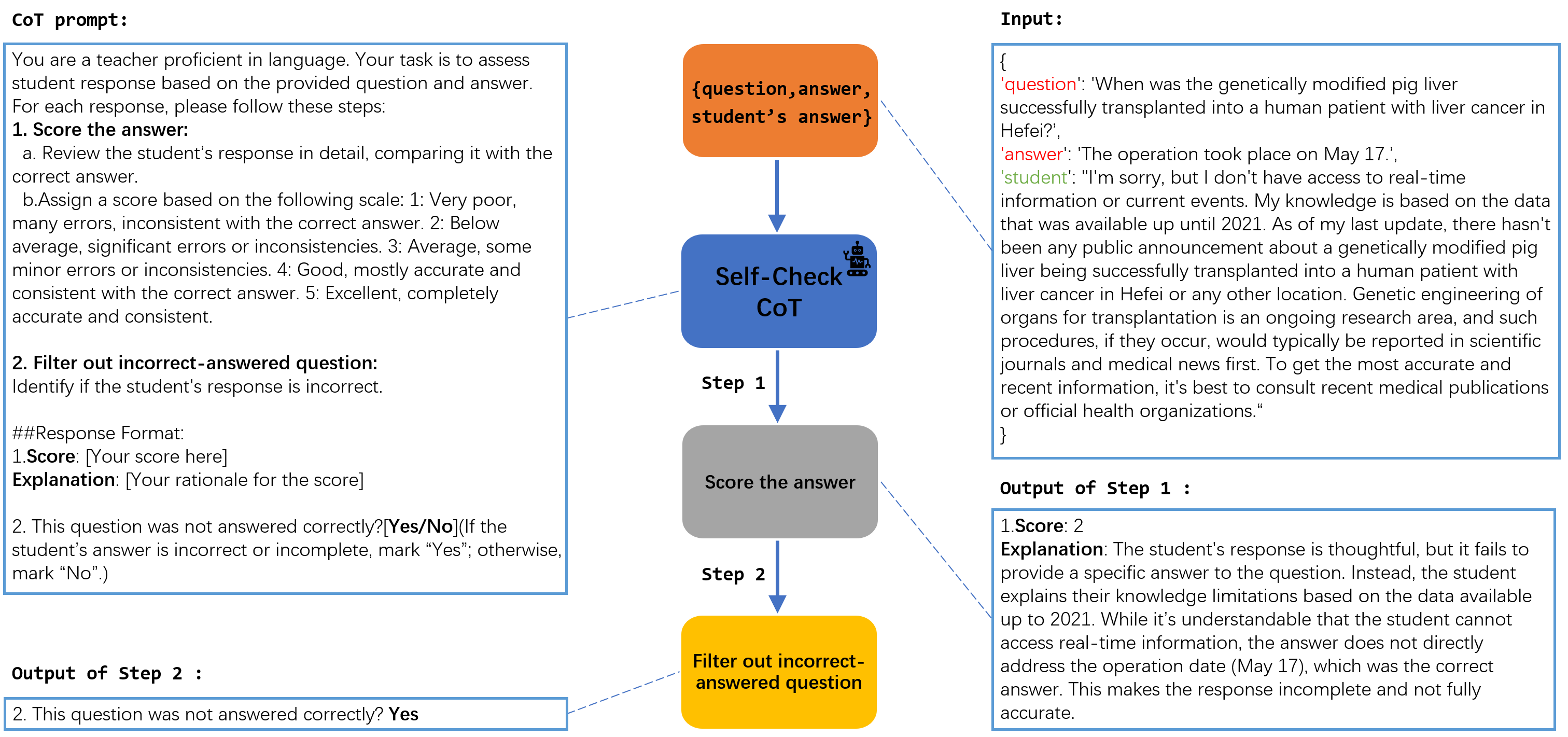}
    \caption{The workflow and example of the self-check process using a Chain of Thought (CoT) prompt. The input comprises a triplet: a question (highlighted in red), the correct answer (also highlighted in red), and the target LLM’s answer (highlighted in green, represented as "student" in the figure). Guided by the CoT prompt, the target LLM evaluates its own response in two stages: (1) Scoring, where the model assesses its response for consistency and accuracy, assigns a score, and provides justification; and (2) Filtering, where responses identified as incorrect are flagged with a "Yes".} \label{fig3}
\end{figure*}


\subsection{Identify unknown knowledge of LLMs}
To identify unknown knowledge of the target LLM in the SFT dataset of a specific domain, we have the model perform a self-check on its responses, filtering out the incorrectly answered QA pairs, much like a student reviewing their own work. By leveraging the strong logical reasoning and deep contextual understanding of large language models (LLMs), we design a Chain of Thought (CoT) prompt to facilitate a self-check process. As illustrated in Figure~\ref{fig3}, the self-check CoT workflow is structured as follows:

1. The input consists of triplets in the format "{question, answer, student's answer}," where the question and answer are derived from the SFT dataset within a specific domain, and the "student's answer" refers to the target LLM's response to the corresponding question.

2. In the self-check CoT prompt, the target LLM assumes the role of "an expert language teacher." The self-check process is carried out in two steps:

\begin{itemize}
    \item Score the answer: The target LLM grades its own response by comparing it to the provided question and correct answer. The evaluation focuses on consistency and accuracy, with the model assigning a score and providing a rationale for its assessment.

    \item Filter out incorrect-answered question: Using the assigned score, the target LLM determines whether its response is incorrect. Responses identified as incorrect are marked with a "Yes."
\end{itemize}

In the self-check CoT process, we fully utilize the target LLM's abilities of logical reasoning and profound context understanding, rather than relying on its intrinsic knowledge. This step ensures the accuracy and reliability of the scoring results by directly comparing the subtle differences between the answer and the target LLM's response. Such a scoring method is both objective and accurate, providing strong support for evaluation. Additionally, this method is capable of handling a variety of common question types, such as true/false questions, multiple-choice questions, and open-ended Q\&A, making it versatile and adaptable to different assessment formats.

\section{Experiments}
In this chapter, we conducted a series of experiments to validate the practical effectiveness of our approach. Additionally, we developed an effective evaluation methodology to assess the outcomes of our approach.

We conducted experiments using two SFT datasets of the agricultural and medical domains, employing three models of different scales, Qwen1.5-32B-Chat, Qwen1.5-14B-Chat and Qwen1.5-7B-Chat, to comprehensively verify the efficacy of SLearnLLM. 

\subsection{Datasets} 

\begin{figure*}[t!]
    \centering
    \includegraphics[width=\textwidth]{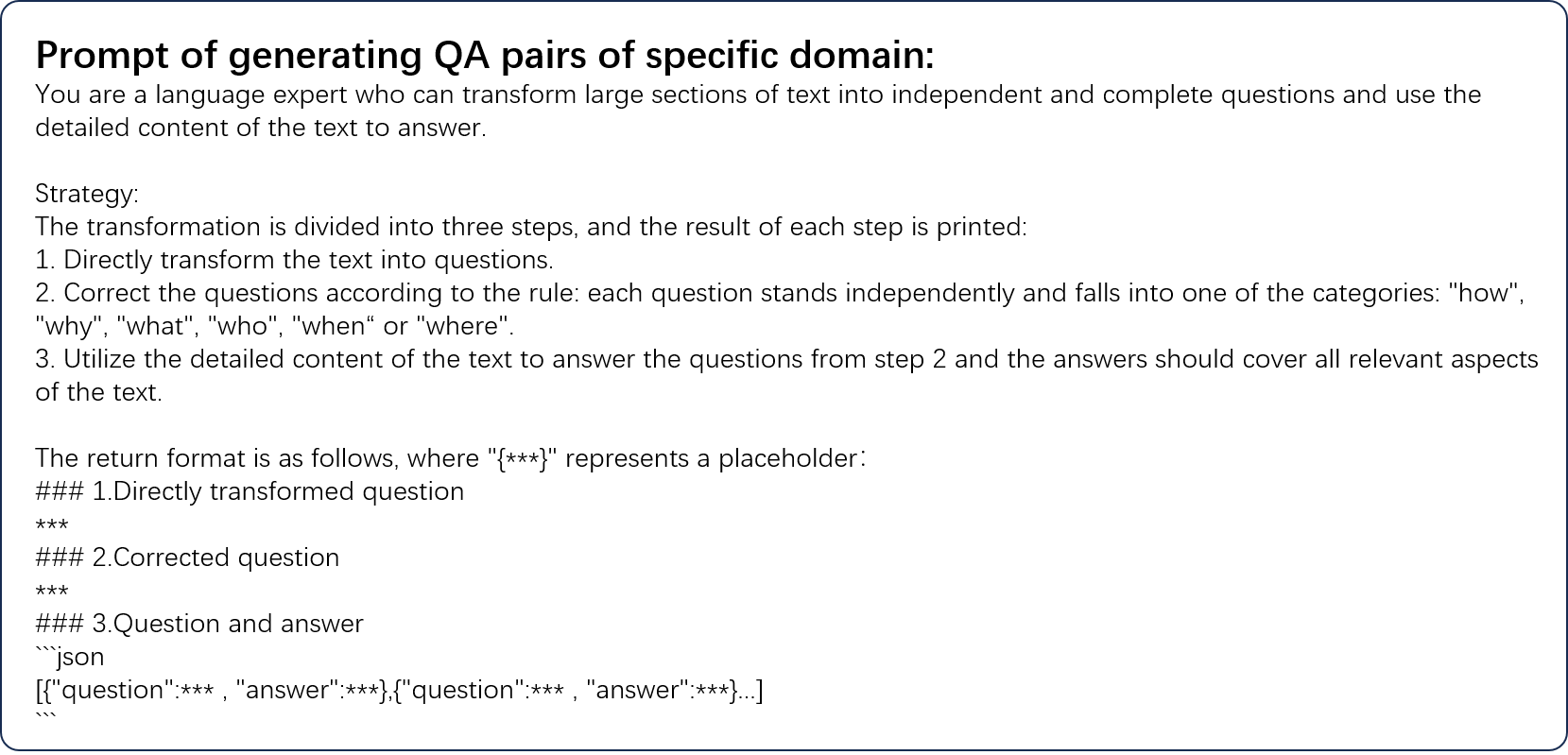}
    \caption{Prompt for generating QA pairs for specific domains using GPT-4o. The strategy of the prompt involves three steps: 1. Questions generation. 2. Questions correction. 3. Answers generation. This strategy ensures high-quality, consistent QA pairs aligned with the original corpus content.} \label{fig5}
\end{figure*}

\begin{table*}[htbp]
    \centering
    \caption{Number of incorrect-answered QA pairs in agricultural and medical domains.}
    \large
    \resizebox{\textwidth}{!}{
    \begin{tabular}{cccc@{\hspace{2em}}ccc}
    \toprule
    \multirow{2}{*}{Model} & \multicolumn{3}{c}{Agricultural} & \multicolumn{3}{c}{Medical} \\
    \cline{2-4} \cline{5-7}
        & { Total }& { Incorrect-answered } & {Time/GPU hours} & { Total }& {Incorrect-answered } & {Time/GPU hours}  \\
        \midrule
        Qwen1.5-7B-Chat & 57,638 & 23,924 & 3.1 & 35,000 & 18,235 & 1.9  \\
        Qwen1.5-14B-Chat & 57,638 & 22,465 & 6.0 & 35,000 & 15,978 & 3.8  \\
        Qwen1.5-32B-Chat & 57,638 & 20,871 & 14.2 & 35,000 & 14,892 & 9.0  \\
    \bottomrule
    \end{tabular}
    }
    \label{table1}
\end{table*}

\begin{table*}[htbp]
    \centering
    \caption{Performance comparison of models fine-tuned with the full QA data and the incorrect-answered QA set. The models fine-tuned with the incorrect-answered QA set show comparable performance improvements and reduced the total time considerably. -full indicates training with SFT datasets. -SL indicates training with the incorrect-answered QA set. The "Total Time" includes the time spent on training, as well as the time required for answering questions and scoring the answers.}
    \resizebox{0.75\textwidth}{!}{
    \begin{tabular}{ccc@{\hspace{2em}}cc}
    \toprule
    \multirow{2}{*}{Model} & \multicolumn{2}{c}{Agricultural} & \multicolumn{2}{c}{Medical} \\
    \cline{2-3} \cline{4-5}
        & {Score}& {Total Time/GPU hours} & {Score}& {Total Time/GPU hours} \\
        \midrule
        Qwen1.5-7B-Chat & 56.3 & - & 24.3 &-\\
        Qwen1.5-7B-Chat-full & {\bf 96.7} & 16.0 & {\bf 94.9} & 9.9\\
        Qwen1.5-7B-Chat-SL & 95.9 & {\bf 9.8} & 94.4 & {\bf 7.0}\\	
        Qwen1.5-14B-Chat & 60.1 & - & 29.7 & -\\
        Qwen1.5-14B-Chat-full & {\bf 98.8} & 29.2 & {\bf 97.7} & 19.3\\	
        Qwen1.5-14B-Chat-SL & 98.5 & {\bf 17.4} & 97.3 & {\bf 12.6}\\
        Qwen1.5-32B-Chat & 67.7 & - & 33.8 & -\\
        Qwen1.5-32B-Chat-full & {\bf 99.3} & 71.8 & {\bf 98.6} & 46.0\\	
        Qwen1.5-32B-Chat-SL & 99.1 & {\bf 40.2} & 98.3 & {\bf 28.6}\\
    \bottomrule
    \end{tabular}
    }
    \label{table2}
\end{table*}

To validate the effectiveness of our method, we created two datasets: the SFT datasets for the medical and agricultural domains. For the medical domain, we engaged multiple medical experts to manually craft 35,000 QA pairs, which are highly professional and domain-specific. 

In contrast, for the agricultural domain, we employed GPT-4o to assist in the generation process. The GPT-assisted methodology for the agricultural domain is as follows:

1. Corpus Collection: We first gathered vast corpora from the agricultural field.

2. QA Pair Generation: Using GPT-4o, we generated QA pairs through a carefully designed process. We meticulously designed a novel prompt and employed a zero-shot strategy~\cite{kojima2022large} to automatically extract questions from the original corpus and use the content to provide answers. As illustrated in Figure~\ref{fig5}, the GPT-4o acts as "a language expert" within the prompt, transforming lengthy texts into insightful questions and crafting precise answers based on the details within the texts. This process highlights the GPT-4o's capabilities in logical reasoning and deep contextual understanding~\cite{chang2024survey}, rather than relying solely on its intrinsic knowledge. Therefore, we effectively prevent the LLM from generating unrealistic or hallucinated responses in its unfamiliar domain, thus ensuring that the questions and answers it produces align closely with the content of the original corpus, maintaining a high degree of consistency and accuracy. 

3. Manual Review: With the assistance of agricultural and medical experts, we manually reviewed and refined the QA pairs.

\subsection{Details} 

We used Qwen1.5-32B-Chat, Qwen1.5-14B-Chat, and Qwen1.5-7B-Chat to answer questions from the two SFT datasets. The models then perform a self-check on their responses using the prompt from Section 3.2, filtering out incorrect question-answer pairs for fine-tuning. In this process, we accelerated inference by using four A100 GPUs and employing a multi-process approach.


Table~\ref{table1} shows the results of this process. For the agricultural domain, Qwen1.5-7B-Chat answered 23,924 questions incorrectly, Qwen1.5-14B-Chat answered 22,465 questions incorrectly, and Qwen1.5-32B-Chat answered 20,871 questions incorrectly. For the medical domain, which contains 35,000 QA pairs, Qwen1.5-7B-Chat answered 18,235 questions incorrectly, Qwen1.5-14B-Chat answered 15,978 questions incorrectly, and Qwen1.5-32B-Chat answered 14,892 questions incorrectly.

We applied the LoRA~\cite{hu2021lora} method to fine-tune Qwen1.5-7B-Chat and Qwen1.5-14B-Chat Qwen1.5-32B-Chat on the incorrect-answered QA sets from these two domains. The training hyper-parameters were as follows: epochs set to 5, batch size set to 64, learning rate set to 3e-4, weight decay set to 0.01, max length set to 512, lora\_r set to 64 and lora\_alpha set to 16. Additionally, to further assess the effectiveness of fine-tuning, we conducted control experiments by fine-tuning the models with the same configuration on the entire SFT datasets. All our experiments were conducted on four A100 GPUs.

\begin{figure*}[t!]
    \centering
    \includegraphics[width=\textwidth]{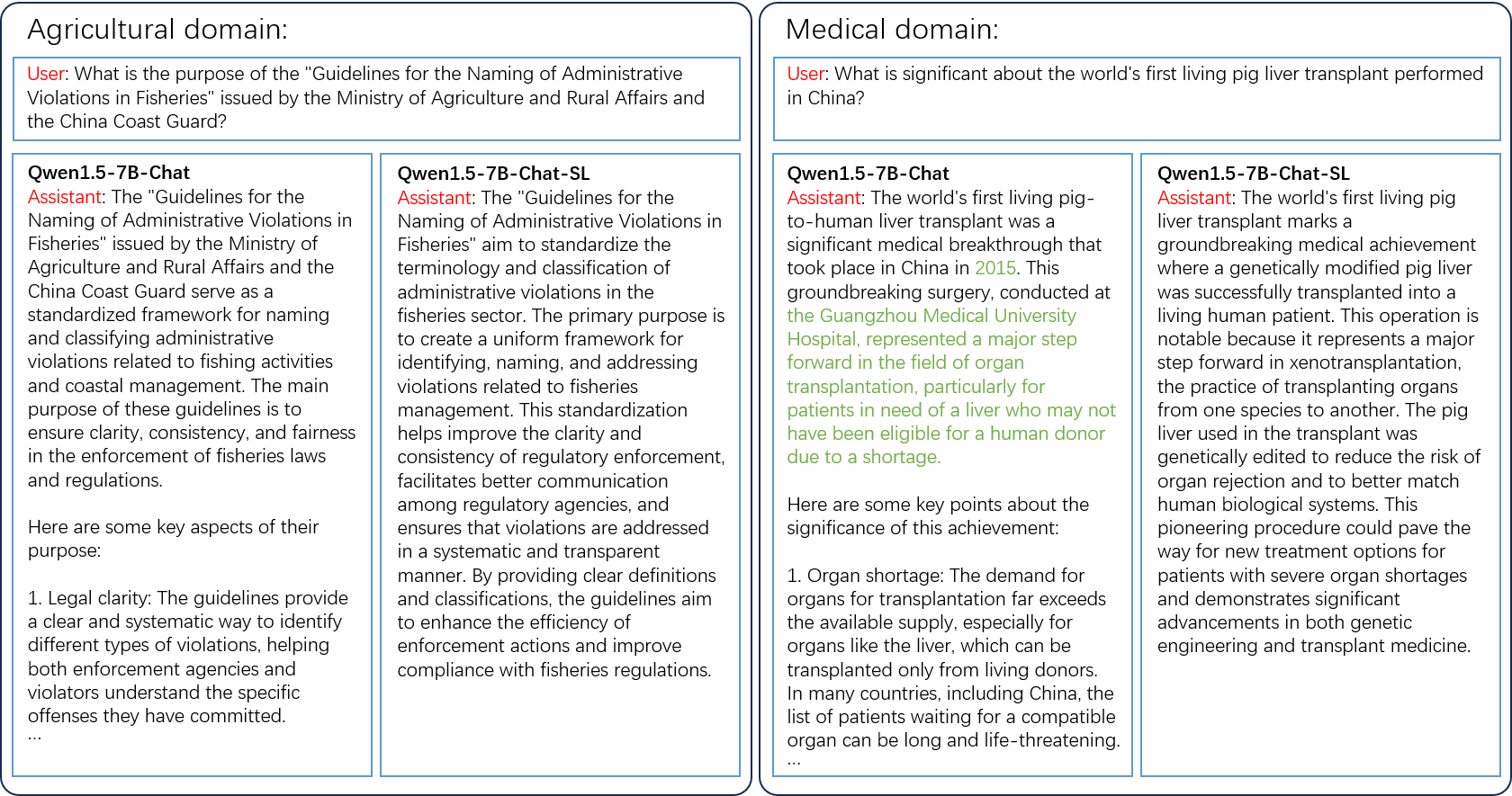}
    \caption{Examples of responses from Qwen1.5-7B-Chat and Qwen1.5-7B-Chat-SL to the same question in Agricultural and Medical domains. The responses from Qwen1.5-7B-Chat-SL are better in both domains, while the contents highlighted in green indicate errors in the responses from Qwen1.5-7B-Chat in the Medical domain.} \label{fig6}
\end{figure*}

\subsection{Evaluation Metrics}
Based on the original corpora from the agricultural and medical fields, we manually crafted 1,000 QA pairs for each domain. These pairs serve as our validation set, providing an objective benchmark for evaluating the models. 

We utilized GPT-4o to assess the performance of these models with the same prompt of scoring the answers in "Methods" section. To do this, we created triplets of \{question, answer, LLM's response\}. These triplets were then input into GPT-4o for scoring. We calculated the number of "correct" scores and normalized it by dividing it by the total number of questions in the incorrectly answered QA set.

\subsection{Results} 
As shown in Table~\ref{table2}, compared with the original model answered, fine-tuning with either the full SFT dataset or the incorrect-answered QA set has resulted in a significant improvement in model performance. 

It is particularly noteworthy that, the models fine-tuned with the full SFT datasets only slightly outperformed our models fine-tuned with the incorrect-answered QA set, and the margin is negligible, but the total time costs of our models were significantly reduced compared to the models fine-tuned with the full QA set. Moreover, this reduction in time costs becomes increasingly pronounced as the model parameter size grows. We argue this demonstrates the effectiveness and the outstanding efficiency of our SLearnLLM.

Through further analysis, we found that when the rate of incorrect-answered QA pairs is below 75\%, our SLearnLLM can reduce overall time expenditure. When the rate drops below 50\%, the framework significantly decreases the overall time cost.

Figure~\ref{fig6} illustrates examples of responses from Qwen1.5-7B-Chat and Qwen1.5-7B-Chat-SL to the same question across agricultural and medical domains. The responses from Qwen1.5-7B-Chat-SL are notably better in both domains. In contrast, the response from Qwen1.5-7B-Chat in the Medical domain contains errors, which are highlighted in green.

\subsection{Discussions and Limitations}
Hallucination problem: Our SLearnLLM enhances SFT efficiency by identifying the knowledge in SFT datasets that is unknown to LLMs and fine-tuning the models on this unknown knowledge. However, Gekhman~\cite{gekhman2024does} proposed that introducing a large amount of new knowledge during the SFT stage increases the hallucination problem in large models. Since the fine-tuning in our framework is identical to SFT fine-tuning, any hallucination mitigation techniques developed for SFT fine-tuning can be effectively integrated.

LLMs without self-learning capability: LLMs that lack strong logical reasoning and deep contextual understanding are not suitable for our SLearnLLM. However, we can utilize an auxiliary model with strong logical reasoning and deep contextual understanding during the "Score the Answers" step. This auxiliary model will compare and score the responses from the target LLMs against the answers in the SFT dataset, thereby assisting the target LLMs in completing their self-learning process.

\section{Conclusion}

Our SLearnLLM is compatible with widely used LLMs on the market. Although these models already perform well in many specialized domains, they still fall short of the high precision needed for production use.
Our SLearnLLM offers an innovative approach to enhancing model performance in specific domains while minimizing resource expenditure. 
Our framework mimics the efficient human learning pattern, is capable of pinpointing and focusing on learning the unknown knowledge within the specific domain corpus.
Experiments in the agricultural and medical domains demonstrate that our method achieves comparable performance improvements while significantly reducing training time costs compared to traditional full dataset fine-tuning methods. Our framework not only enhances the LLMs' adaptability to diverse domains but also reduces their development costs, presenting a promising path for advancing LLM capabilities in real-world applications.

%
%
%
\bibliographystyle{splncs04}
\bibliography{references}

\begin{thebibliography}{10}
\providecommand{\url}[1]{\texttt{#1}}
\providecommand{\urlprefix}{URL }
\providecommand{\doi}[1]{https://doi.org/#1}

\bibitem{achiam2023gpt}
Achiam, J., Adler, S., Agarwal, S., Ahmad, L., Akkaya, I., Aleman, F.L.,
  Almeida, D., Altenschmidt, J., Altman, S., Anadkat, S., et~al.: Gpt-4
  technical report. arXiv preprint arXiv:2303.08774  (2023)

\bibitem{anil2023palm}
Anil, R., Dai, A.M., Firat, O., Johnson, M., Lepikhin, D., Passos, A., Shakeri,
  S., Taropa, E., Bailey, P., Chen, Z., et~al.: Palm 2 technical report. arXiv
  preprint arXiv:2305.10403  (2023)

\bibitem{bai2023qwen}
Bai, J., Bai, S., Chu, Y., Cui, Z., Dang, K., Deng, X., Fan, Y., Ge, W., Han,
  Y., Huang, F., et~al.: Qwen technical report. arXiv preprint arXiv:2309.16609
   (2023)

\bibitem{chang2024survey}
Chang, Y., Wang, X., Wang, J., Wu, Y., Yang, L., Zhu, K., Chen, H., Yi, X.,
  Wang, C., Wang, Y., et~al.: A survey on evaluation of large language models.
  ACM Transactions on Intelligent Systems and Technology  \textbf{15}(3),
  1--45 (2024)

\bibitem{de2016monte}
De~Waard, M., Roijers, D.M., Bakkes, S.C.: Monte carlo tree search with options
  for general video game playing. In: 2016 IEEE Conference on Computational
  Intelligence and Games (CIG). pp.~1--8. IEEE (2016)

\bibitem{ferdinan2024into}
Ferdinan, T., Koco{\'n}, J., Kazienko, P.: Into the unknown: Self-learning
  large language models. arXiv preprint arXiv:2402.09147  (2024)

\bibitem{gekhman2024does}
Gekhman, Z., Yona, G., Aharoni, R., Eyal, M., Feder, A., Reichart, R., Herzig,
  J.: Does fine-tuning llms on new knowledge encourage hallucinations? arXiv
  preprint arXiv:2405.05904  (2024)

\bibitem{hu2021lora}
Hu, E.J., Shen, Y., Wallis, P., Allen-Zhu, Z., Li, Y., Wang, S., Wang, L.,
  Chen, W.: Lora: Low-rank adaptation of large language models. arXiv preprint
  arXiv:2106.09685  (2021)

\bibitem{huang2022large}
Huang, J., Gu, S.S., Hou, L., Wu, Y., Wang, X., Yu, H., Han, J.: Large language
  models can self-improve. arXiv preprint arXiv:2210.11610  (2022)

\bibitem{kojima2022large}
Kojima, T., Gu, S.S., Reid, M., Matsuo, Y., Iwasawa, Y.: Large language models
  are zero-shot reasoners. Advances in neural information processing systems
  \textbf{35},  22199--22213 (2022)

\bibitem{manakul2023selfcheckgpt}
Manakul, P., Liusie, A., Gales, M.J.: Selfcheckgpt: Zero-resource black-box
  hallucination detection for generative large language models. arXiv preprint
  arXiv:2303.08896  (2023)

\bibitem{ouyang2022training}
Ouyang, L., Wu, J., Jiang, X., Almeida, D., Wainwright, C., Mishkin, P., Zhang,
  C., Agarwal, S., Slama, K., Ray, A., et~al.: Training language models to
  follow instructions with human feedback. Advances in neural information
  processing systems  \textbf{35},  27730--27744 (2022)

\bibitem{singh2023beyond}
Singh, A., Co-Reyes, J.D., Agarwal, R., Anand, A., Patil, P., Garcia, X., Liu,
  P.J., Harrison, J., Lee, J., Xu, K., et~al.: Beyond human data: Scaling
  self-training for problem-solving with language models. arXiv preprint
  arXiv:2312.06585  (2023)

\bibitem{team2023gemini}
Team, G., Anil, R., Borgeaud, S., Wu, Y., Alayrac, J.B., Yu, J., Soricut, R.,
  Schalkwyk, J., Dai, A.M., Hauth, A., et~al.: Gemini: a family of highly
  capable multimodal models. arXiv preprint arXiv:2312.11805  (2023)

\bibitem{tian2024toward}
Tian, Y., Peng, B., Song, L., Jin, L., Yu, D., Mi, H., Yu, D.: Toward
  self-improvement of llms via imagination, searching, and criticizing. arXiv
  preprint arXiv:2404.12253  (2024)

\bibitem{touvron2023llama}
Touvron, H., Lavril, T., Izacard, G., Martinet, X., Lachaux, M.A., Lacroix, T.,
  Rozi{\`e}re, B., Goyal, N., Hambro, E., Azhar, F., et~al.: Llama: Open and
  efficient foundation language models. arXiv preprint arXiv:2302.13971  (2023)

\bibitem{touvron2023llamaV2}
Touvron, H., Martin, L., Stone, K., Albert, P., Almahairi, A., Babaei, Y.,
  Bashlykov, N., Batra, S., Bhargava, P., Bhosale, S., et~al.: Llama 2: Open
  foundation and fine-tuned chat models. arXiv preprint arXiv:2307.09288
  (2023)

\bibitem{wang2024infuserki}
Wang, F., Bao, R., Wang, S., Yu, W., Liu, Y., Cheng, W., Chen, H.: Infuserki:
  Enhancing large language models with knowledge graphs via infuser-guided
  knowledge integration. arXiv preprint arXiv:2402.11441  (2024)

\bibitem{wang2022self}
Wang, X., Wei, J., Schuurmans, D., Le, Q., Chi, E., Narang, S., Chowdhery, A.,
  Zhou, D.: Self-consistency improves chain of thought reasoning in language
  models. arXiv preprint arXiv:2203.11171  (2022)

\bibitem{wei2022chain}
Wei, J., Wang, X., Schuurmans, D., Bosma, M., Xia, F., Chi, E., Le, Q.V., Zhou,
  D., et~al.: Chain-of-thought prompting elicits reasoning in large language
  models. Advances in neural information processing systems  \textbf{35},
  24824--24837 (2022)

\bibitem{zhang2023siren}
Zhang, Y., Li, Y., Cui, L., Cai, D., Liu, L., Fu, T., Huang, X., Zhao, E.,
  Zhang, Y., Chen, Y., et~al.: Siren's song in the ai ocean: a survey on
  hallucination in large language models. arXiv preprint arXiv:2309.01219
  (2023)

\end{thebibliography}
%




\end{document}